\documentclass[
11pt,              
letterpaper,       
fleqn,
headnosepline,	   
oneside,           
onecolumn,         
openright,         
noonelinecaption,
cleardoubleempty,
smallheadings]
{scrartcl}         
\addtokomafont{caption}{\sffamily\footnotesize}
\setkomafont{pagenumber}{\sffamily\footnotesize}
\addtokomafont{pagenumber}{\sffamily\footnotesize}
\setkomafont{captionlabel}{\sffamily\bfseries\footnotesize}
\setcapindent{0em} 

\usepackage{makecell}
\usepackage[usestackEOL]{stackengine}
\usepackage{graphicx}
\usepackage{psfrag}
\usepackage{amsmath}
\usepackage{amssymb}
\usepackage{color}
\usepackage{xcolor}
\usepackage{rotate}
\usepackage{makeidx}
\usepackage{cite}
\usepackage{eufrak}
\usepackage{gensymb}
\usepackage[varumlaut]{yfonts}
\usepackage{mleftright}
\usepackage{helvet}   
\usepackage{mathpazo}
\normalfont
\usepackage[T1]{fontenc}
\usepackage{textcomp}
\usepackage{longtable} 
\usepackage[only,llbracket,rrbracket]{stmaryrd}
\usepackage{bm}
\begin{document}
\newcommand{\beq} {\begin{equation}}
\newcommand{\eeq} {\end{equation}}
\newcommand{\D}   {\displaystyle}
\newcommand{\divg}{\mbox{\rm{div}}\,}
\newcommand{\clearemptydoublepage}{\newpage{\pagestyle{empty}\cleardoublepage}}
\newcommand{\Divg}{\mbox{\rm{Div}}\,}
\newtheorem{remark}      {\bf{\sffamily{Remark}}}
\newtheorem{definition}  {\bf{\sffamily{Definition}}}
\renewcommand{\sc}{}
\renewcommand{\Psi}{\psi}
\renewcommand{\varrho}{\vartheta}
\renewcommand{\arraystretch}{1.3}
\sloppy
\def\sca   #1{\mbox{\rm #1}{}}
\def\mat   #1{\mbox{\bf #1}{}}
\def\vec   #1{\mbox{\boldmath $#1$}{}}
\def\ten   #1{\mbox{\boldmath $#1$}{}}
\def\scas  #1{\mbox{{\scriptsize{${\rm{#1}}$}}}{}}
\def\vecs  #1{\mbox{{\boldmath{\scriptsize{$#1$}}}}{}}
\def\tens  #1{\mbox{{\boldmath{\scriptsize{$#1$}}}}{}}
\def\up    #1{^{\mbox{\rm{\footnotesize{#1}}}}}
\def\down  #1{_{\mbox{\rm{\footnotesize{#1}}}}}
\def\ltr   #1{\mbox{\sf{#1}}}
\def\bltr  #1{\mbox{\sffamily{\bfseries{{#1}}}}}
\vspace*{0.8cm}
\begin{center}
{\sffamily\bfseries\Large{Minimal activation with maximal reach:}}\\[4pt]
{\sffamily\bfseries\Large{Reachability clouds of bio-inspired slender manipulators}}\\
\vspace*{1.0cm}
Bartosz Kaczmarski$^*$,
Derek E. Moulton$^{**}$,
Alain Goriely$^{**}$,
Ellen Kuhl$^*$\\ [6.pt]
$^*$ Department of Mechanical Engineering \\
Stanford University, Stanford, California, United States \\[4.pt]
$^{**}$ Mathematical Institute \\
University of Oxford, Oxford, OX2 6GG, United Kingdom\\
\end{center}
\vspace*{0.9cm}
{\sffamily{\bfseries{Abstract.}}}
In the field of soft robotics, 
flexibility, adaptability, and functionality 
define a new era of robotic systems 
that can safely
deform, reach, and grasp. 
To optimize the design of soft robotic systems, 
it is critical to understand 
their configuration space and full range of motion 
across a wide variety of design parameters.
Here we integrate 
extreme mechanics and soft robotics 
to provide quantitative insights 
into the design 
of bio-inspired soft slender manipulators 
using the concept of reachability clouds.
For a minimal three-actuator design 
inspired by the elephant trunk, 
we establish an efficient and robust reduced-order method
to generate reachability clouds 
of almost half a million points each
to visualize the accessible workspace 
of a wide variety of manipulator designs.
We generate an atlas of 256 reachability clouds 
by systematically varying the key design parameters 
including the 
fiber count, revolution, tapering angle, and activation magnitude.
Our results demonstrate 
that reachability clouds 
not only offer an immediately clear perspective 
into the inverse problem of control, 
but also introduce powerful metrics to characterize 
reachable volumes, unreachable regions, and actuator redundancy to quantify the performance of soft slender robots.
Our study provides new insights 
into the design of soft robotic systems 
with minimal activation and maximal reach
with potential applications in 
medical robotics, 
flexible manufacturing, and the
autonomous exploration of space.  

\vspace*{0.5cm}

\clearpage
\section{Motivation}
\label{sec:intro}
Throughout the past decade, the field of soft robotics has revolutionized the landscape of robotic systems with unprecedented levels of {\it{flexibility}}, {\it{adaptability}}, and {\it{functionality}} \cite{coyleBioinspiredSoftRobotics2018,rusDesignFabricationControl2015b,amjadiStretchableSkinMountableWearable2016,kimSoftRoboticsBioinspired2013}.
Soft robotic systems deform and adjust flexibly:
they can safely grasp objects of different size and shape,
operate functionally, mimic natural organisms or biological processes, and
respond adaptively to environmental changes
\cite{eshaghiDesignManufacturingApplications2021,caoUntetheredSoftRobot2018,guoBioinspiredMultimodalSoft2022}.
With broad applications in
medical robotics, 
flexible manufacturing, and
autonomous exploration,  
a critical element to 
optimize their design 
is the comprehensive visualization 
of the full range 
of possible movements and configurations.
This is exactly the purpose of {\it{reachability clouds}}.
Reachability clouds are
discrete point clouds 
that graphically illustrate 
the reachable workspace of a robotic system; 
they provide valuable insights 
into the capabilities and limitations of a robot 
including 
workspace analysis, 
motion planning, and, most importantly, 
design optimization \cite{guptaNatureRobotWorkspace1986,zachariasCapturingRobotWorkspace2007,trivediSoftRoboticsBiological2008b,huang2023,strohle2022}.   
However, creating meaningful reachability clouds of millions of points can be computationally expensive, especially when the actuation is {\it{continuous}} and deformations are {\it{finite}}. \\[6.pt]
We have previously proposed a computationally inexpensive method to efficiently and robustly calculate the actuated configurations of soft slender manipulators \cite{kaczmarskiActiveFilamentsCurvature2022c}.
Inspired by the intricate musculature of the elephant trunk \cite{leanza2024}, we inform our theory of active soft slender structures by the extreme mechanics of thin elastic rods \cite{betsch2002,betsch2003,oreillyModelingNonlinearProblems2017,millerBucklingThinElastic2015,moultonMorphoelasticRodsIII2020e,johanns2024}, 
which are enjoying increasing popularity in soft-robotic research \cite{dellasantinaModelBasedControlSoft2023,jonesBubbleCastingSoft2021,kratchmanGuidingElasticRods2017,tillRealtimeDynamicsSoft2019,huang2022}.
Here we adapt our method to create multiple reachability clouds throughout the design space of a soft slender manipulator to maximize its reachable workspace and enhance its overall performance.
Specifically, we focus on the four critical design parameters: the number of fibers, fiber revolution, tapering angle, and activation magnitude.
Using a three-fiber minimal design archetype with one longitudinal and two helical fibers of opposite handedness \cite{kaczmarskiMinimalDesignElephant2024}, we compute a total of $16 \times 16$ reachability clouds of 400 thousand points each for the 256  design variations. We quantify the performance of each design in terms of its {\it{concave and convex cloud volumes}} and its {\it{unreachable volume fraction}}.
We compare the minimal three-actuator design against a redundant four-actuator design with two longitudinal and two helical fibers in terms of their reachability clouds and mean activation-space distances.
Critical to our study is the computational efficiency of our model \cite{kaczmarskiActiveFilamentsCurvature2022c,kaczmarskiSimulationToolPhysicsInformed2023c,kaczmarskiBayesianDesignOptimization2023b} that allows us to generate reachability clouds of millions of points at a computational cost of only a few minutes on a standard desktop computer.\\[6.pt]
\begin{figure*}[t!]
    \centering
    \includegraphics[width = \textwidth]{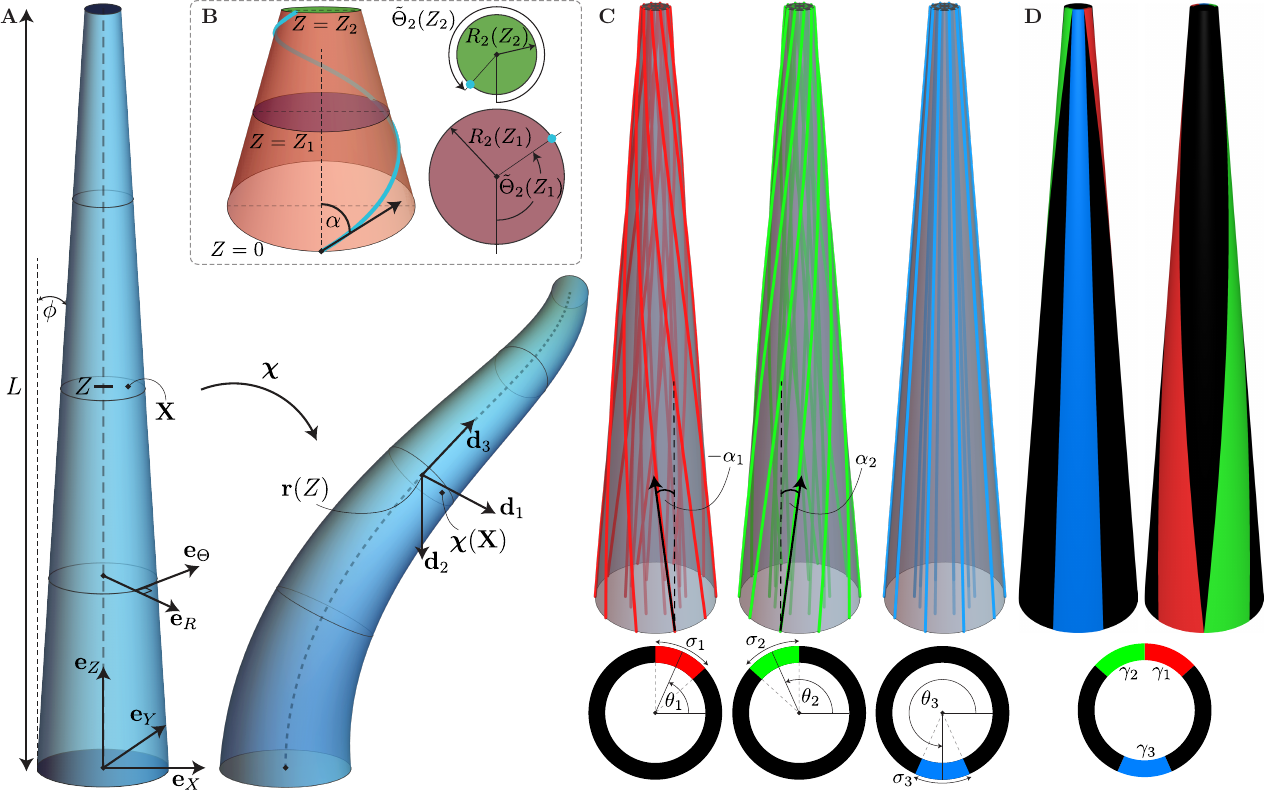}
    \caption{{\bf{\sffamily{Analytical model and minimal design of bio-inspired slender manipulator.}}} (a) Summary of continuum mechanics quantities pertinent to the reduced-order active filament theory \cite{kaczmarskiActiveFilamentsCurvature2022c}. A one-dimensional initial configuration deforms through a map $\bm{\chi}$ to an activated configuration with a centerline $\mathbf{r}$ and directors $\mathbf{d}_i$. (b) Helical fiber revolution around a tapered structure (left), with two cross sections (right) extracted at two different $Z$ values. (c) An example of three fiber architectures with $n_1=n_2=n_3=1$ activatable bundles in each architecture (top), and an example cross-sectional placement of the bundles (bottom). (d) Front and back view of the manipulator design resulting from fiber architectures and bundle placements in (c), which constructs a tapered minimal design \cite{kaczmarskiMinimalDesignElephant2024} with activations $\gamma_1$, $\gamma_2$, $\gamma_3$ in the three actuators.}
    \label{fig:model}
\end{figure*}
Throughout this manuscript, we systematically explore and compare  reachability clouds for varying combinations of design parameters to derive several exciting trends and general design guidelines for soft robotic systems. We interpret the intricate atlas of reachability clouds in the context of design efficacy, quantify the reachable volumes, and establish  trade-offs between reachability and configuration versatility. Through the juxtaposition with a redundant four-fiber design, we highlight the inherent benefits of the three-fiber actuation minimality.

\section{Analytical model}\label{sec:model}
We apply the active filament theory to model the mechanics of soft slender manipulators with contractile fibers of arbitrary architecture \cite{kaczmarskiActiveFilamentsCurvature2022c}. 
Following a general continuum mechanics formulation of the deformation $\bm{\chi}$ of a tubular structure, we perform a dimensional reduction to express the deformation as 
\begin{align}\label{eq:chi}
\bm{\chi} = \bm{r} + \sum_{i = 1}^3 \varepsilon \, e_i \, \mathbf{d}_i,
\end{align}
where $\varepsilon$ is the ratio of the characteristic cross-sectional radius to the length $L$ of the manipulator, $\varepsilon \, e_i$ are the cross-sectional reactive strains, $\mathbf{r}$ is the centerline curve of the slender structure, and $\mathbf{d}_i$ with $i \in\{1,2,3\}$, defines the orthonormal director basis that describes the orientation of the cross section along the centerline, see Fig.\ \ref{fig:model}A
\cite{Goriely2017,moultonMorphoelasticRodsIII2020e}. 
We write the kinematics of the resulting one-dimensional representation of the structure as
\begin{equation}
\mathbf{r}'   = \zeta \, \mathbf{d}_3 \qquad \; 
\mathbf{d}_i' = \zeta \, \mathbf{u}\times \mathbf{d}_i \, ,
\end{equation}
where 
$(\circ)'$ denotes the derivative with respect to the longitudinal coordinate $Z$ of the initial configuration, 
$\zeta$ is the axial extension, and 
$\mathbf{u} = \sum_{i=1}^3\mathsf{u}_i \, \mathbf{d}_i$ is the Darboux vector of curvatures \cite{moultonMorphoelasticRodsPart2013g}. For given boundary conditions for $\mathbf{r}$ and $\mathbf{d}_i$, the functions $\mathsf{u}_i$ fully define the deformed configuration of the manipulator in this reduced-order setting.
To express the curvatures $\mathsf{u}_i$ for arbitrary fiber architectures and inhomogeneous contractile activations, we multiplicatively decompose the deformation gradient, 
$\mathbf{F} = \text{Grad} \, {\bm\chi} = \mathbf{A} \cdot \mathbf{G}$, 
into an elastic part $\mathbf{A}$ and an activation part $\mathbf{G}$ 
\cite{kaczmarskiActiveFilamentsCurvature2022c}. 
Critically, the specific form of the deformation $\bm{\chi}$ in eq.\ \eqref{eq:chi} enables an analytical energy minimization that results in  explicit expressions for the curvatures~$\mathbf{u}$. 
We assume a tapered truncated cone geometry of the manipulator with a varying cross-sectional radius $R_2(Z)$, a homogeneous Young's modulus, and $M$ helical fiber architectures embedded in a single tubular ring with inner radius $R_1(Z)$ and outer radius $R_2(Z)$ 
\cite{kaczmarskiActiveFilamentsCurvature2022c}, 
and express the post-activation axial extension as
\begin{equation}
  \hat{\zeta} 
= 1 + \frac{1}{4 \, R_2^2}
  \sum_{i = 1}^{M}
  \delta_{0,i} \, a_{0,i},
\end{equation}
and the related curvatures as
\begin{equation}
\begin{array}{l@{\hspace*{0.1cm}}c@{\hspace*{0.1cm}}c@{\hspace*{0.0cm}}l}
    \displaystyle{\hat{\mathsf{u}}_1} 
&=& \displaystyle{-\frac{2}{3R_2^4}}
&   \displaystyle{\sum_{i = 1}^{M}\delta_{1,i} \, A_i \sin\left(\varphi_i - \tilde{\Theta}_{2,i}\right)}\\
    \displaystyle{\hat{\mathsf{u}}_2} 
&=& \displaystyle{-\frac{2}{3R_2^4}}
&   \displaystyle{\sum_{i = 1}^{M}\delta_{2,i} \, A_i \cos\left(\varphi_i - \tilde{\Theta}_{2,i}\right)}\\
    \displaystyle{\hat{\mathsf{u}}_3} 
&=& \displaystyle{\frac{2(1 +\nu)}{3R_2^4}}
&   \displaystyle{\sum_{i = 1}^{M}\delta_{3,i} \, a_{0,i}} \, ,
\end{array}
\end{equation}
where $\delta_{j,i}(Z)$ are functions that depend on the helical angle $\alpha_i$ of the $i$-th architecture, the domain geometry, and the Poisson's ratio $\nu$; see Supplementary Material. 
Fig.\ \ref{fig:model}C, top, shows an example of three fiber architectures with $\alpha_1<0$, $\alpha_2>0$, and $\alpha_3 = 0$. The quantities $A_i=(a_{1,i}^2 + b_{1,i}^2)^{1/2}$, $\varphi_i = -\arctan(b_{1,i}, a_{1,i})$, and $a_{0,i}$ describe the fibrillar activation, 
\begin{equation}
\begin{split}
a_{0,i} &= \frac{\sigma_i}{\pi}\sum_{j = 1}^{n_i}\gamma_{j,i}\\
a_{1,i} &= \frac{2\sin(\sigma_i/2)}{\pi}\sum_{j = 0}^{n_i - 1}\gamma_{j + 1,i} \cos \, \left(\theta_{0,i} + \frac{2 \pi j}{n_i}\right)
\\
b_{1,i} &= \frac{2\sin(\sigma_i/2)}{\pi}\sum_{j = 0}^{n_i - 1}\gamma_{j + 1,i} \sin \, \left(\theta_{0,i} + \frac{2 \pi j}{n_i}\right) \, ,
\end{split}
\end{equation}
where, in the $i$-th architecture, $n_i$ is the number of independent fiber bundles that are equidistant in the cross-sectional polar angle $\theta$, $\gamma_{j,i}$ denotes the fibrillar activation value in the $j$-th bundle, $\sigma_i$ is the angular extent of all fiber bundles, and $\theta_{0,i}$ is the cross-sectional angular offset of all bundles, see Fig.\ \ref{fig:model}C, bottom. Finally, for a helical angle $\alpha_i$, the function $\tilde{\Theta}_{2,i}(Z)$ gives the total fiber rotation in the interval $[0, Z]$ on the outer boundary $R = R_2$ of the manipulator,
\begin{equation}
\tilde{\Theta}_{2,i}(Z) = -\left(\frac{\tan \alpha_i}{\sin \phi}\right) \log\left(\frac{R_2(Z)}{R_2(0)}\right) \, ,
\end{equation}
where $\phi$ is the tapering angle of the truncated cone geometry,
see Supplementary Material for treating the singularity at $\phi = 0$.
Fig.\ \ref{fig:model}B provides a visual description of $\tilde{\Theta}_2$. 

In a \textit{minimal design} inspired by the elephant trunk's muscle anatomy \cite{leanza2024, kaczmarskiMinimalDesignElephant2024}, three distinct architectures are present: (1) left-handed helical with $\alpha_1 < 0$, (2) right-handed helical with $\alpha_2 >0$, and (3) longitudinal with $\alpha_3 = 0$. For minimality, the three corresponding angular extents $\sigma_i = 48\degree$ are all equal, and each architecture contains one controllable fiber bundle such that $n_i = 1$, and the activation simplifies to $\gamma_{1,i} = \gamma_i$, $i \in \{1,2,3\}$. Inspired by the elephant trunk, the two helical architectures are symmetric, i.e., $\alpha_1 = -\alpha_2$, and the angular offsets are $\theta_{0,1} = 66\degree$, $\theta_{0,2} = 114\degree$, and $\theta_{0,3} = 270\degree$, resulting in a symmetric design, see Fig.\ \ref{fig:model}D. Finally, we set $R_2(0) = L / 16$, $R_1(0) = L/12$, and $\nu = 1/2$ for incompressibility.  

\section{Reachability cloud atlas}
\label{sec:atlas}
We use the analytical model to compute the deformed configurations of a family of fiber-based manipulator designs. By randomly sampling $N$ activations $\gamma_i$ in each fiber bundle over some feasible set $\gamma_i\in[\gamma_{\text{min}}, \gamma_{\text{max}}]$, we can rapidly sample the configuration space of a manipulator as $N\rightarrow \infty$. In fact, the low computational cost of the model allows $N$ to be as large as $2 \cdot 10^6$, at a computational cost of only several minutes on a standard desktop computer. Due to high sampling density, $N\sim 10^6$ generates a sufficiently exhaustive mapping from the activation space to the configuration space. To visualize and analyze this mapping, we use the concept of {\it{reachability clouds}}, a set of all $N$ end-effector positions of the manipulator resulting from all the $N$ computed configurations.\\[6.pt]
\begin{figure*}[t!]
    \centering
    \includegraphics[width = \textwidth]{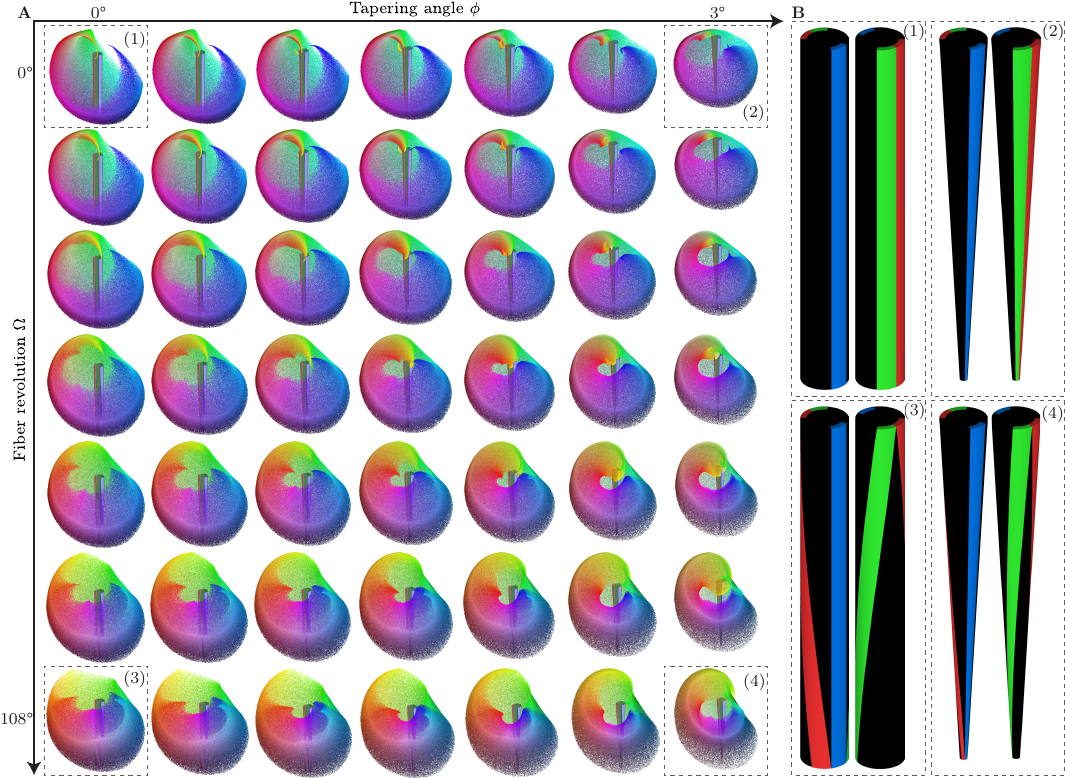}
    \caption{{\bf{\sffamily{Reachability cloud atlas for varying fiber revolutions and tapering angles.}}} (a) Atlas of
    $7\times 7$ reachability clouds 
    with 400,000 activation samples each, generated using ranges of $\Omega\in[0\degree, 108\degree]$ for the helical fiber revolution and $\phi\in[0\degree, 3\degree]$ for the tapering angle. We use an RGB color scale to color-code the cloud points, where the activation magnitudes $|\gamma_{1,i}|\in[0, 1.67]$ in the red, green, and blue fiber bundles contribute to the R, G, and B components of the RGB point color. (b) Opposing views of the four designs of the four corner cases in the atlas: (1) no fiber helicity and no tapering, (2) no fiber helicity and maximal tapering, (3) maximal fiber helicity and no tapering, (4) maximal fiber helicity and tapering.}
    \label{fig:atlas}
\end{figure*}
While computing a reachability cloud for a given design provides information about its control capabilities, it is even more insightful to investigate how the reachability cloud changes in response to adjustments of the design. Our previous studies revealed that fiber helicity has a significant effect on reachability and that introducing tapering can yield more intricate configurations \cite{kaczmarskiMinimalDesignElephant2024}. Therefore, we created a {\it{reachability cloud atlas}}, a set of 256 reachability clouds that explores the effects and interplay of fiber helicity $\Omega = \tilde{\Theta}_2(L)$ and tapering angles $\phi$ through a $16\times16$ sweep of these two parameters with fiber activations sampled from $\gamma_i \in[-5/3, 0]$. Fig.\ \ref{fig:atlas}A shows a $7\times 7$ reduced visualization of this reachability cloud atlas. Fig.\ \ref{fig:atlas}B shows the corner-case fiber designs corresponding to the clouds at $\Omega = 0\degree$, $108\degree$ and \mbox{$\phi = 0\degree$, $3\degree$}. \\[6.pt]
All designs of this atlas are members of our {\it{minimal bio-inspired design}} family with two symmetrically arranged helical fibers of opposite handedness and a longitudinal fiber at the opposite side of the $Z = 0$ cross section \cite{kaczmarskiMinimalDesignElephant2024}. At the maximal revolution of $\Omega = 108\degree$, the helical fibers meet the longitudinal fiber at the distal end of the manipulator, and the maximal tapering angle of $\phi = 3\degree$ yields approximately a 6.2-fold decrease in the outer radius of the structure at the distal end. All designs are symmetric about the plane passing through the center of the cross section and the midpoint axis of the longitudinal fiber.\\[6.pt]
We make several interesting observations about the morphology of the reachability clouds as a function of the two design parameters: First, increasing either the fiber helicity or the tapering angle expands the reachable space of the manipulator in the directions normal to the plane of symmetry. That is, the clouds of designs with low $\Omega$ and $\phi$ are significantly more localized around the plane of symmetry. This observation is consistent with the intuition that the torsion generated by fiber helicity facilitates motion out of the symmetry plane. Second, larger fiber revolutions promote an increase in the overall thickness of the cloud. In particular, while the clouds for $\Omega = 0\degree$ are closely representable by two-dimensional surfaces, the $\Omega = 108\degree$ clouds have a more volumetric appearance, as evidenced by their more distant cloud boundaries. Third, the morphologies of clouds generated with higher tapering angles are generally more intricate, with an extended region of reachability behind, i.e., on the helical-fiber side, and above the manipulator. This phenomenon likely occurs due to the inherent ability of tapered designs to generate variable curvature and torsion, even with only a single fiber. Further, tapered designs produce significant curling in their deformed configurations which is, in part, responsible for the morphological intricacies of the tapered-design clouds.

\section{Concave and convex cloud volumes}
\label{sec:volumes}
\begin{figure*}[t!]
    \centering
    \includegraphics[width = \textwidth]{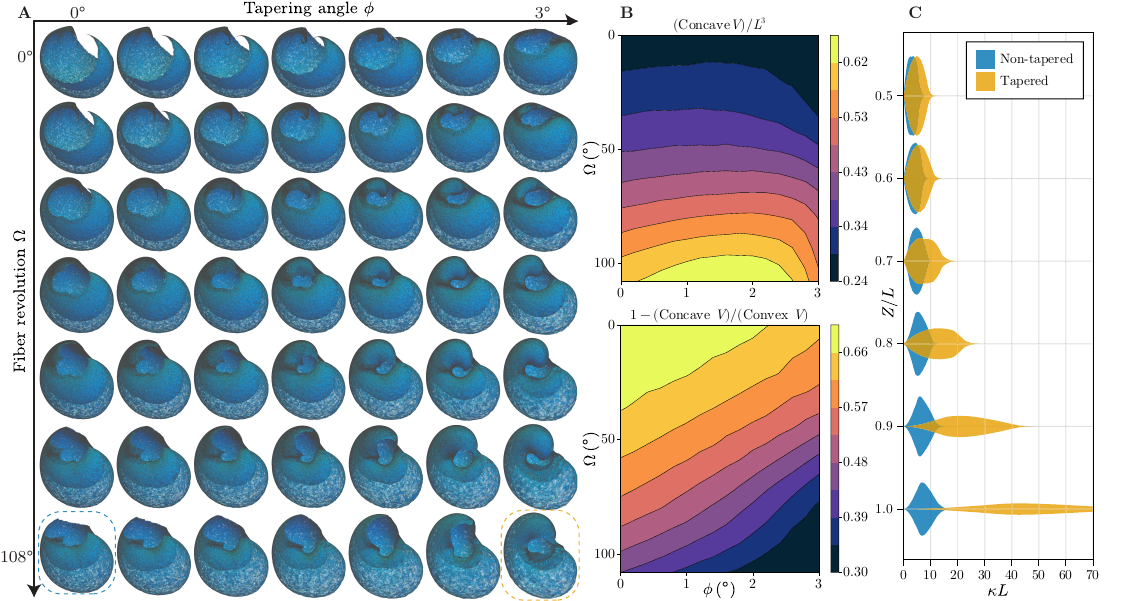}
    \caption{{\bf{\sffamily{Reachability cloud volumes for varying fiber revolutions and tapering angles.}}} (a) Tight concave hull mesh boundaries for all $7\times7$ reachability clouds in the atlas. (b) Concave hull volume normalized by $L^3$ and plotted as a function of the fiber revolution and tapering angles (top). Volume fraction of the unreachable region inside the convex hull plotted as a function of the fiber revolution and tapering angles (bottom). We generated both contour plots for the full atlas of $16 \times 16$ clouds. (c) Comparison of configuration versatility between a non-tapered design, marked blue in (a), and a tapered design, marked orange in (a). We plot the distribution of curvatures $\kappa L$ at 6 uniformly-spaced points in the distal half of the structure, and show that the configuration space of the tapered design is richer than that of the non-tapered design.}
    \label{fig:volumes}
\end{figure*}

From the perspective of control, the {\it{volume of a reachability cloud}} is a powerful scalar metric to characterize the versatility of a manipulator design. To quantify design versatility across our  design space, we compute the cloud volumes across our reachability atlas. Specifically, we use the $\alpha$-shapes method \cite{edelsbrunnerShapeSetPoints1983a} to create a tight three-dimensional {\it{concave hull}} mesh, which accounts for the geometric complexity of the cloud boundary. Figure\ \ref{fig:volumes}A visualizes the concave hulls for all $7 \times 7$ clouds in Fig.\ \ref{fig:atlas}A. Based on visual inspection, all concave hulls accurately represent the overall geometry of the clouds, with only minor mesh construction artifacts in highly tapered designs with high fiber helicity. We conclude that the volume inside the concave mesh is a good approximation of the cloud volume. \\[6.pt]
We report the normalized volumes of the clouds as a function of the design parameters in the contour plot in Fig.\ \ref{fig:volumes}B, top. Notably, the reachability cloud volume generally increases with increasing fiber helicity. This result is consistent with our observation from Fig.\ \ref{fig:atlas}A that the reachability clouds become thicker and span a larger space away from the symmetry plane for larger fiber revolutions. We also observe that there exists an optimal tapering angle $\phi^* = 2\degree$ for which the normalized volume is maximal with the value of $V / L^3 \approx 0.67$. The precise location of this maximum generally depends on the design parameters. Larger tapering angles reduce the cloud volume: Overly curled configurations sacrifice the ability of the end-effector to reach distant points, as the curling mechanism brings the distal tip closer to the main body of the manipulator. \\[6.pt]
In addition to the cloud volume itself, it is insightful to quantify the manipulator's effectiveness to explore its surrounding space. Figure\ \ref{fig:volumes}B, bottom, illustrates the relative proportion of the cloud's {\it{convex hull}} volume occupied by unreachable space. We define the {\it{unreachability}} metric as the unreachable volume fraction within the convex hull,
$
 \text{UNR}
= 1 - V_{\rm{conc}} / V_{\rm{conv}}
$, 
where 
$V_{\rm{conc}}$ and $V_{\rm{conv}}$
denote the volumes of the concave and convex hulls.
The smaller the value of $\text{UNR}$, 
the closer the concave and convex hull volumes, and
the more effective the corresponding design at navigating the surrounding space without encountering unreachable regions. \mbox{Fig.\ \ref{fig:volumes}B}, bottom, shows that $\text{UNR}$ decreases when either the fiber helicity or the tapering angle increases, which suggests an inherent advantage of incorporating these two design elements for more effective manipulator control. The monotonous relationship between the $\text{UNR}$ metric and $\Omega$ and $\phi$ suggests that there might be benefits to creating highly tapered designs, even at the expense of a decrease in the cloud volume indicated by Fig.\ \ref{fig:volumes}B, top.\\[6.pt]
While the analyzed clouds are a useful tool to quantify the reachability properties for a given design's end-effector, they do not describe the {\it{manipulator configurations}} that give rise to each end-effector position that makes up a point in the cloud. In practical terms, understanding the configurations that effectively produce a given reachability cloud is critical for a variety of control tasks such as obstacle avoidance and navigating restricting environments. To better understand and compare the configuration spaces of the designs in the atlas, we use statistical analysis. In Fig.\ \ref{fig:volumes}C, we show the distributions of curvatures corresponding to the 400,000 configurations 
for a non-tapered design 
with $\phi = 0\degree$ in blue and 
for a tapered design 
with $\phi = 3\degree$ in yellow,
at 6 uniformly-spaced points in the distal half of each design. Evidently, the tapered design produces increasingly higher curvatures closer to its tip, which is consistent with the bending stiffness decreasing due to tapering. Strikingly, not only is the peak of each curvature distribution shifted, but the range of curvatures is also significantly widened. The tapered design can generate most of the curvatures in the non-tapered design by choosing appropriate activation triplets, while the non-tapered design cannot reproduce the large range of curvatures accessible to the tapered design. This effect is more pronounced in points closer to the tip of the manipulator and explains the significance of tapering in nature, such as an elephant trunk {\it{wrapping}} around a distant branch of a tree.

\section{Redundancy}
\label{sec:redundancy}

\begin{figure*}[t!]
    \centering
    \includegraphics[width = \textwidth]{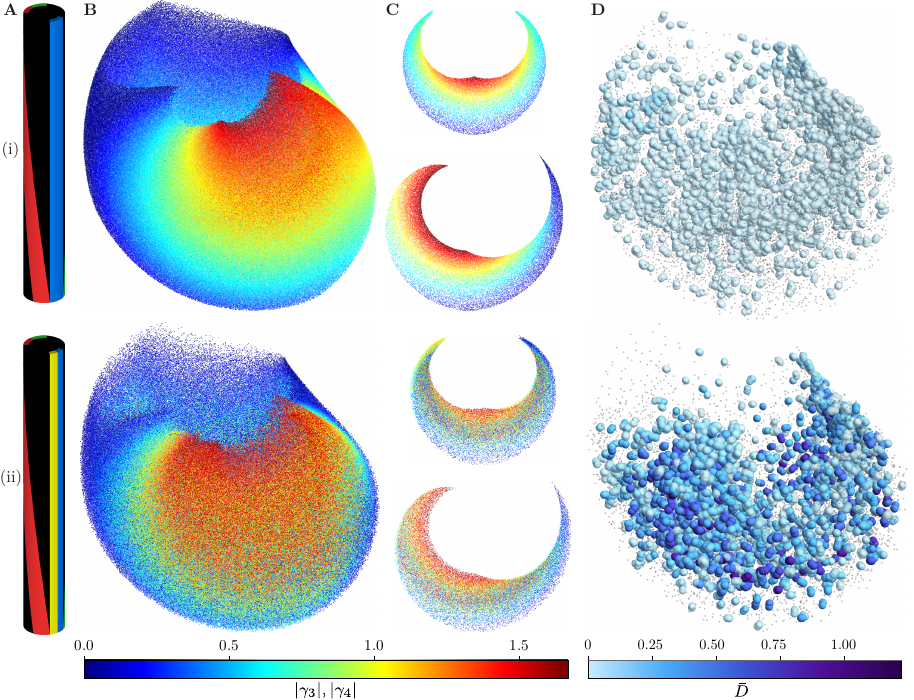}
    \caption{{\bf{\sffamily{Reachability clouds of 
    the minimal and redundant designs.}}} (a) Arrangement of three and four independently actuated fibers in the minimal design, top, and redundant design, bottom. Both designs contain two helical fibers, red and green; the minimal design includes one longitudinal fiber, blue, the redundant design includes two longitudinal fibers, blue and yellow. (b) Reachability clouds of the two designs color-coded by the activation magnitude $|\gamma_{3}|$ in the blue fiber for the minimal design, and $|\gamma_{4}|$ in the yellow fiber in the redundant design. Each cloud consists of 2 million points. (c) Slices of the corresponding reachability clouds in (b). All slices pass through the origin and, for each cloud, the top slice is in the frontal plane, while the bottom slice is in the right-facing plane. (d) Mean activation-space distance visualized over a 10,000-point subset of each design's reachability cloud for $r_\mathcal{S} = L / 60$. Regions with large mean activation distances indicate a significant effect of redundancy caused by large activation changes within small sub-volumes of the cloud. Light gray points depict spheres that only bound the center point and no other cloud points.}
    \label{fig:redundancy}
\end{figure*}

Throughout our analysis of the atlas, we explored the reachability properties under the assumption of {\it{design minimality}} where a {\it{unique activation triplet}} generates each configuration in the workspace. For three independently actuated fibers, the reachability cloud is a three-dimensional volume. Instead, for two fibers only, the reachable geometry reduces to a two-dimensional surface, and for one fiber, it becomes a one-dimensional curve. Since the positions of the end-effector are inherently three-dimensional, adding a fourth independently contractible fiber introduces a {\it{redundant degree of freedom}}. As a result, {\it{several activation quadruplets}} can achieve the same end-effector position. Our reachability clouds can provide immediate quantitative insight into the effects of design redundancy.\\[6.pt]
We introduce redundancy by splitting the longitudinal fiber of the minimal design, in blue, into two independent longitudinal fibers, in blue and yellow, in addition to the two helical fibers, see Fig.\ \ref{fig:redundancy}A. Figure\ \ref{fig:redundancy}B visualizes reachability clouds with $2\cdot 10^6$ points each for both the minimal design, top, and redundant design, bottom, color-coded by the activation magnitudes of the blue or yellow fibers. Importantly, the reachability cloud of the minimal design displays a continuous and single-valued progression of longitudinal fiber activation across the entire volume. In contrast, the reachability cloud of the redundant design appears more noisy due to the overlap of points with different fiber activation values. The slices of the redundant design in Fig.\ \ref{fig:redundancy}C confirm the multi-valued nature of the mapping from the end-effector positions to the underlying fiber activations. Furthermore, the end-point density in the redundant design cloud is more heterogeneous than that of the mimimal design cloud; the region in the front and to the sides of the two longitudinal fibers exhibits a higher point density in comparison to the opposite side of the manipulator. As such, the redundant design shows a localization effect in that two neighboring longitudinal fibers contribute a higher degree of redundancy on the front side of the manipulator.\\[6.pt]
To gain further insight into the non-uniqueness of activation for redundant designs, we introduce a mean distance metric that characterizes the variation of activation in a small neighborhood of points of the reachability cloud. We define this {\it{mean activation-space distance}} as
$
 \bar{D}(p_i, r_\mathcal{S}) 
=\sum_{j=1}^{K} 
 \|\bm{\Gamma}_{i,j}^\mathcal{S} - \bm{\Gamma}_i\|_2 / K \,
$,
where $p_i\in\mathbb{R}^3$ is the $i$-th point in the cloud, $i\in\{1,\ldots, N\}$, $\bm{\Gamma}_i\in \mathbb{R}^M$ is a vector of fibrillar activation values for point $p_i$, and $\bm{\Gamma}_{i,j}^\mathcal{S}$ is the activation vector at one of the $K$ points inside the boundary of a sphere of radius $r_\mathcal{S}$ centered at $p_i$. As a result, $\bar{D}$ computes the mean Euclidean distance in the $M$-dimensional activation space within a physical three-dimensional neighborhood of each point in the cloud. \\[6.pt]
Figure\ \ref{fig:redundancy}D shows the mean activation-space distances plotted over a 10,000-point subset of the complete cloud for both the minimal and redundant designs. For the minimal design, the values of $\bar{D}$ are generally small and  uniformly distributed throughout the entire cloud, which is indicative of the continuous changes in activation during cloud traversal. In contrast, for the redundant design, $\bar{D}$ varies significantly throughout the cloud and assumes much larger values in some regions. For our example, mean distance values on the order of $1.0$ are highly indicative of redundancy. \\[6.pt]
Regional variations in the mean activation-space distance $\bar{D}$ coincide with our previous observation of regional variations in sparsity throughout the reachability cloud. The points in the front and to the sides of the two longitudinal fibers indeed demonstrate a more significant effect of redundancy compared to the points on the opposite side of the design. We conclude that we can use the mean activation-space distance $\bar{D}$ as a quantitative lens into the localized effects of soft actuator redundancy. \\[6.pt]
From a high-level perspective, the single-valued mapping from the workspace to the activation space of the minimal design is a noteworthy manifestation of how the generally ill-posed inverse problem in an infinite-degree-of-freedom soft-robotic manipulator becomes well-posed when discretely activated through three fiber bundles. In fact, the number $M = 4$ of independent fibers marks the boundary of ill-posedness of the inverse problem. However, we emphasize that, since redundancy is typically contextualized within a given motion task, not every design considered redundant for a given task results in the ill-posedness of the inverse problem. In our analysis, we considered redundancy in the context of the most exhaustive set of deformations permitted by a prescribed range of activation magnitudes.

\section{Conclusions}
\label{sec:conclusions}

Our study presents a comprehensive investigation of the {\it{design and reachability}} of bio-inspired soft slender manipulators through the generation and analysis of reachability clouds. We established a highly efficient computational framework to explore the influence of critical design parameters—fiber count, revolution, tapering angle, and activation magnitude—on the manipulator's workspace. The creation of reachability clouds allowed us to visualize and quantify the convoluted workspaces of minimal and redundant actuator designs.\\[6.pt]
Using this method, we found that both fiber helicity and tapering are pivotal in expanding the reachability of soft manipulators, enabling them to access a larger volume of space with intricate maneuvers that mimic the adaptive and flexible movements we can observe in nature. Both parameters can significantly change design functionality, as they govern the trade-off between reachability, design complexity, and control precision.\\[6.pt]
Our work is inspired on a key idea: the reduction of complex three-dimensional slender structures to one-dimensional curves that allow extremely fast computation and rapid generation of millions of configurations. This computational tool allows us to generate enough points to accurately compute the concave and convex hulls of the reachability clouds, from which we have established guidelines for maximizing workspace while maintaining control efficacy. The statistical examination of the configuration space provides quantitative insights into its richness, and emphasizes the benefits of tapering for generating diverse configurations.\\[6.pt]
Exploring design redundancy reveals its double-edged nature: while increasing redundancy increases the manipulator's flexibility, it also introduces complexity in the control scheme due to the presence of multiple configurations that can reach the same endpoint. Specifically, the introduction of redundancy in actuator design, through the addition of a fourth independently contractible fiber, illustrates the  balance between increased configuration versatility and the challenges in control strategy due to non-unique activation tuples. Therefore, our study highlights the importance of a balanced approach to design, where the benefits of increased reachability and configuration diversity must be weighed against the added complexity in control. More generally, our study shows how designs with a finite number of controllable degrees of freedom in the form of discrete fibers can help address the well-known ill-posedness of the inverse problem in an otherwise infinite-degree-of-freedom soft-robotic system. In contrast to the four-fiber design, we demonstrate that three-fiber designs offer a unique activation solution to the inverse problem for every point in their workspace. \\[6.pt]
The big-data nature of our reachability atlas and the well-posedness of the inverse problem for our three-fiber design suggest to expand the use of data-driven approaches for effective design and control of fiber-based manipulators. Future work could explore the fully dynamic system---rather than the quasi-static deformations computed here---to examine the effects of transient deformations on the reachability clouds. Beyond the minimal three-fiber design of this study, a natural expansion would be to explore reachability atlases for more complex fiber architectures.\\[6.pt]
Taken together, our study integrates extreme mechanics and soft robotics to provide quantitative insights into the design of bio-inspired soft slender manipulators using the concept of reachability clouds. 
We demonstrate that reachability clouds not only offer an immediately clear perspective into the inverse problem of reachability, but also introduce powerful metrics to characterize reachable volumes, unreachability, and  redundancy, all of which quantify the performance of soft slender robots. 
As such, this work lays the theoretical and computational foundations for automated design, control, and optimization of soft robotic systems. 
We expect that our systematic quantitative study of soft slender manipulators will generalize to other soft robots and guide the design of more effective soft robotic systems with the objective of minimal activation and maximal reach. 
\appendix
\section*{Appendix A. Expressions for $\delta_0$, $\delta_1$, $\delta_2$, $\delta_3$}
\noindent
According to the theory in \cite{kaczmarskiActiveFilamentsCurvature2022c}, the functions $\delta_{0}(Z)$, $\delta_{1}(Z)$, $\delta_{2}(Z)$, and $\delta_{3}(Z)$ read 
\begin{align}
&\delta_0 = 2 \left(R_1^2-R_2^2\right) \nu -\frac{2 (1+\nu )}{c_{\phi }^2-c_{\alpha}^2}{ \log
   \left(\frac{\left(1+R_1^2 c_{\phi }^2\right) \left(1+R_2^2 c_{\alpha}^2\right)}{\left(1+R_2^2 c_{\phi }^2\right) \left(1+R_1^2 c_{\alpha}^2\right)}\right)}\\
\begin{split}
&\delta_1  = \frac{2}{c_{\phi } c_{\alpha}
   \left(c_{\phi }^2-c_{\alpha}^2\right)}\bigg[3 (1+\nu )
   \cdot\left(\arctan\left(R_1 c_{\phi }\right)
   -\arctan\left(R_2 c_{\phi }\right)\right) c_{\alpha}
   +\left(R_1^3-R_2^3\right) \nu  c_{\phi }^3 c_{\alpha}\\[-10.pt]
&\hspace*{2.65cm}-c_{\phi }
   \Big(3 (1+\nu ) \left(\arctan\left(R_1 c_{\alpha}\right)-\arctan\left(R_2 c_{\alpha}\right)\right)
   +\left(R_1^3-R_2^3\right) \nu
    c_{\alpha}^3\Big)\bigg] = \delta_2 
\end{split}\\
&\delta_3 = \frac{3}{c_{\phi }^2 c_{\alpha}^2 \sqrt{c_{\phi }^2-c_{\alpha}^2}} \bigg[
   \left(T(-R_1)+T(R_1) - T(-R_2) - T(R_2)\right) c_{\phi }^2
   +6 c_{\alpha}
   \left(-S(R_1)+S(R_2)\right)\bigg],
\end{align}
where
\begin{equation}
T(R) = \arctan\mleft(\frac{c_\alpha + ic_\phi^2 R}{S(R)}\mright) \qquad
S(R) = \sqrt{\mleft(c_\phi^2 - c_\alpha^2\mright)\mleft(1 + c_\phi^2 R^2\mright)} \qquad
c_\phi = \frac{\tan \phi}{R_2} \qquad
c_\alpha = \frac{\tan\alpha}{R_2}.
\end{equation}
Note that the $\delta_3$ equation contains the imaginary unit $i$ to simplify the expression; it is still true that $\delta_3 \in\mathbb{R}$. For a manipulator with $M$ distinct fiber architectures, we obtain the corresponding expressions $\delta_{0,k}$, $\delta_{1,k}$, $\delta_{2,k}$, $\delta_{3,k}$ for the $k$-th architecture, $k\in{1,\ldots, M}$, by simply substituting $\alpha \leftarrow \alpha_k$ in each of the expressions above.

\section*{Appendix B. Fiber rotation $\tilde{\Theta}_2$}
\noindent 
We stated that, based on \cite{kaczmarskiActiveFilamentsCurvature2022c}, the total helical fiber rotation at $R = R_2$ in the $i$-th architecture throughout the interval $[0, Z]$ and for a linearly tapered geometry is
\begin{align}
\tilde{\Theta}_{2,i}(Z) = -\mleft(\frac{\tan \alpha_i}{\sin \phi}\mright) \log\mleft(\frac{R_2(Z)}{R_2(0)}\mright),
\end{align}
which contains a singularity for the non-tapered case $\phi = 0$. We recover the case of the rotation $\tilde{\Theta}_{2,i}$ of a given fiber $\mathcal{F}$ for a non-tapered geometry by evaluating the general expression from \cite{kaczmarskiActiveFilamentsCurvature2022c}:
\begin{align}
\tilde{\Theta}_{2,i}(Z) = \tan\alpha_i \int_{0}^Z \frac{\sqrt{1 + (R^0 f'(s))^2}}{R_2(0)f(s)}\,ds,
\end{align}
where $R^0$ is the distance from the center of the cross-section to the fiber $\mathcal{F}$ at $Z = 0$, and $f(Z)$ is the normalized tapering profile such that $R_2(Z) = R_2(0) f(Z)$. The non-tapered case corresponds to $f(Z) = 1$, $Z \in[0,L]$, which gives the simple expression
\begin{align}
\tilde{\Theta}_{2,i}(Z) = \frac{Z}{R_2} \tan\alpha_i.
\end{align}

\section*{Appendix C. Helical angles $\alpha_i$}
\noindent
For tapered domains, the fiber architecture is also appropriately tapered to prevent premature termination of fibers. As follows from \cite{kaczmarskiActiveFilamentsCurvature2022c}, we make the following remarks regarding the helical angle definition $\alpha_i$ pertinent to the treatment of non-tapered and tapered geometries:
\begin{itemize}
    \item Given a cross-section center $O$ and a point $P$ at $R = R_2$ from $O$, for non-tapered geometries ($\phi = 0$), the helical angle $\alpha_i$ is exactly the angle between the fiber direction $\mathbf{m}$ at a point $P$ and the tangent vector $\mathbf{e}_\Theta = \mathbf{e}_Z \times \overrightarrow{OP}$.
    \item For tapered geometries with $\phi \neq 0$, $\alpha_i$ does \textit{not} denote the angle defined in (a). Instead, it defines the angle between the projection of the fiber direction $\mathbf{m}$ onto a plane normal to $\mathbf{e}_R$ at $P$ and the vector $\mathbf{e}_\Theta = \mathbf{e}_Z \times \overrightarrow{OP}$.
    \item As follows from (a) and (b), throughout the text, $\alpha_i$ refers to an angle definition at a distance $R = R_2$ from $O$. 
    \item In general, the helical angle that either $\mathbf{m}$ or the projection of $\mathbf{m}$ makes with $\mathbf{e}_\Theta = \mathbf{e}_Z \times \overrightarrow{OP'}$ changes as the distance $R\in[R_1,R_2]$ of a point $P'$ from $O$ changes. However, all expressions in \cite{kaczmarskiActiveFilamentsCurvature2022c} adapted here only depend on the angle at $R = R_2$.
\end{itemize}
All remarks above refer to the fibers arranged in the initial configuration before fibrillar activation.

\section*{Acknowledgments}
\noindent
This work was supported 
by the NSF CMMI Award 2318188 Mechanics of Bioinspired Soft Slender Actuators to Ellen Kuhl.

\bibliographystyle{elsarticle-num} 
\begin{small}
\bibliography{references}
\end{small}





\end{document}